# DEEP-LEARNING-BASED COMPUTER VISION APPROACH FOR THE SEGMENTATION OF BALL DELIVERIES AND TRACKING IN CRICKET

Kumail Abbas[1], Muhammad Saeed[1], M. Imad Khan[2], Khandakar Ahmed[2,*], Hua Wang[2]

[1]Department of Computer Science, The University of Karachi, Karachi

[2]Intelligent Technology Innovation Lab (ITIL), Victoria University, Melbourne

{ KumailAbb@Gmail.com , Saeed@uok.edu.pk, theimad@gmail.com, khandakar.ahmed@vu.edu.au, hua.wang@vu.edu.au}

## Abstract

There has been a significant increase in the adoption of technology in cricket recently. This trend has created the problem of duplicate work being done in similar computer vision-based research works. Our research tries to solve one of these problems by segmenting ball deliveries in a cricket broadcast using deep learning models, MobileNet and YOLO, thus enabling researchers to use our work as a dataset for their research. The output from our research can be used by cricket coaches and players to analyze ball deliveries which are played during the match. This paper presents an approach to segment and extract video shots in which only the ball is being delivered. The video shots are a series of continuous frames that make up the whole scene of the video. Object detection models are applied to reach a high level of accuracy in terms of correctly extracting video shots. The proof of concept for building large datasets of video shots for ball deliveries is proposed which paves the way for further processing on those shots for the extraction of semantics. Ball tracking in these video shots is also done using a separate RetinaNet model as a sample of the usefulness of the proposed dataset. The position on the cricket pitch where the ball lands is also extracted by tracking the ball along the y-axis. The video shot is then classified as a full-pitched, good-length or short-pitched delivery.

## Keyword

# 1. Introduction

Cricket is arguably one of the most popular sports in the world. It is particularly popular in South-Asian countries, where it is the most played sport. Unlike football, hockey, basketball and other sports where the game is played continuously, progress in a cricket match is made on a ball-by-ball basis. The key moment in cricket is when the bowler delivers a ball and the batsman plays the shot. A total of six deliveries are bowled in an over.

Technology has already been adopted in cricket at the international level and to some extent at the domestic level. Most of this technology is based on computer vision techniques. To broadcast a cricket match, different types of cameras are used which are assembled on the cricket ground in different positions. These include static cameras as well as moving cameras such as spider cams. In our work, we propose a model using deep learning models to extract video shots from the fixed camera containing the video shot of the delivery of the ball from behind the bowler hereafter referred to as the front camera. There are numerous hurdles in extracting this video shot from the broadcast video, which include changes in brightness, movement of the cameras, occlusion, replays, advertisements, etc. In addition to the inherent challenges, there are also other problems associated with broadcast videos such as the difficulty in distinguishing the pitch and the ground, especially when more than one pitch is visible from the front camera. A lot of research has been conducted to overcome these problems using special cameras with markers and fixed positioning. The main problem with the existing work is that special cameras must be fitted in particular positions which makes this research difficult to implement.

The related research discussed in this paper includes tracking the movement of players and balls and extracting semantics from these videos. This involves processing video shots from the front camera as well as tracking the trajectory of the ball and generating highlights of the match, etc. Several works have been conducted in which front view video shots are extracted but only one (Gupta & Sakthi Balan, 2019) is related to the area which we address in our research paper. The results of this research are not taken into consideration due to their low accuracy. On the contrary, we present a model with substantial reliability which is easy to adopt as it is completely based on a broadcast video of a cricket match. The main objective of our research is to address the lack of an available dataset for such video shots. This dataset will enable researchers to apply computer vision and video processing techniques directly to these video shots. Several research works have been conducted where these video shots are extracted manually by the researcher (Kumar et al., 2014). The video shots extracted by our model are also essential from a coaching perspective to help coaches analyze and improve the ball delivery from the bowler and the stroke play from the batsman.

## 1.1 Segmentation of Ball Deliveries

The most readily available footage of international and domestic matches have at least one camera set up in front of the batsman which is used to view the bowler running to deliver the ball. This makes our work applicable to large datasets, hence opening the window to commercialization in our research at a later stage. There are a very small number of commercial products available where players can set up the whole pitch with lots of different cameras, sensors, and even carpeted pitch so that they can analyze themselves but these require a lot of capital and have constraints including a large setup and immobility. Our research also rectifies this by only relying on the video footage and eliminating any need for sensors and extra setup. The advancements in deep learning have enabled us to perform this task simply by processing videos already available.

The task of extracting the front run-up of the bowler and the delivery of the ball using computer vision techniques require many small problems to be solved first, including analyzing the brightness and contrast of different videos, eliminating visual noise, frame alignment based on different broadcast videos which sometimes requires markers to be placed on the pitch in particular positions, etc., thus inducing special requirements for video processing.

All of the aforementioned problems require a large amount of work to be conducted by applying different computer vision techniques. When applying a deep learning-based approach, these problems are solved automatically due to the way the deep learning-based models are built. Object detection in deep learning is not affected by mild changes to the contrast and brightness of videos and the built-in visual noise. The frame alignment and angles are also countered by training frames rigorously at different angles. Hence, the advancements in deep learning models enable us to achieve high performance and accuracy.

Another problem that is encountered in extracting videos from the broadcast video is the extraction of video shots from the primary video source. This problem is less likely to be solved using the built-in deep learning nature and requires core computer vision-based methods as well. The video frames of a video are continuous and are temporally semantic. Exploiting this feature of video frames, there are plenty of methods available which we implemented including analyzing the absolute difference in the color component of the continuous frames and applying object detection using deep learning-based models but the most reliable method comes out to be background subtractor which gives you background mask of the image.

The extracted video shots also need to be categorized as either a live delivery or a replay of one of the previous deliveries. In broadcast videos, the same ball can be replayed several times to viewers so our model must differentiate between a live delivery and a replay of the delivery. To address this problem, we analyzed the speed of the movement of objects in frames using the macroblock motion and bitrate information that is readily accessible from the MPEG video with very minimal decoding (Kobla et al., 1999). We also detected the scorecard at the bottom of the broadcast video using different computer vision techniques. Scorecard detection has been shown to be the most reliable method to identify replays in broadcast videos.

We develop a generic model which takes broadcast videos as the input and outputs small clips of video shots in which the bowler delivers the ball and the batsman plays the shot. This is illustrated in Figure 1.

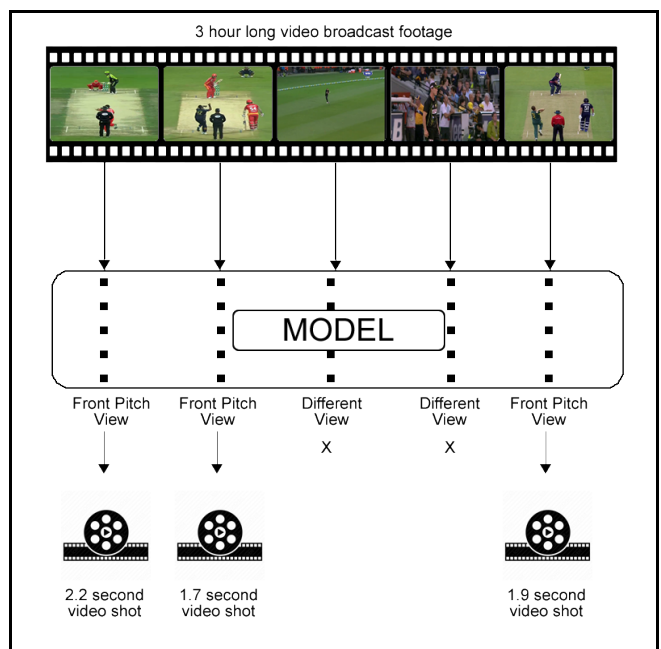

Figure 1. Flow of segmentation of video shots containing ball deliveries from a larger video

## 1.2 Challenges

There are several challenges in segmenting Front Pitch View (FPV) video shot from the broadcast videos. The primary challenge is to differentiate the video segments of the bowler run up with the video segments when the broadcast is done from the zoomed out front camera which shows the whole pitch along with the players. These video segments are displayed mostly immediately after the shot is played by the batter. The examples of these are shown in Figure 2.

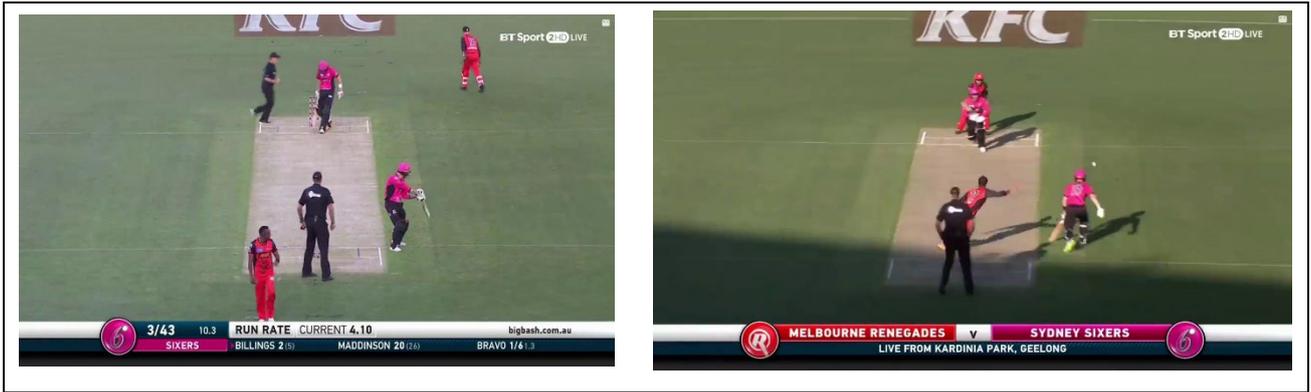

Figure 2. Scenes similar to Front Pitch View (FPV)

Another major challenge is to detect the boundary of the video segment after the batter has played the shot when the ball is followed on by the same camera. This occurs mostly when the ball goes behind the batter. There is no CUT or FADE and the ball is directly followed even after the ball touches the bat. As the ball is continuously focused and followed, the background subtractor also does not work well in this case.

As far as ball tracking is concerned, the frames sequence in broadcast video represents a 2D image. This does not contain the depth to determine how far the ball has traveled towards the batter. We solved this problem by mapping pitch scale on actual pitch and then determining the zoom factor using object detection on the batter which will be discussed later.

## 1.3 Ball Tracking and Classification

After extracting the video shots of the ball delivery, the ball is tracked from the moment it is released from the bowler's hands until it pitches on the ground to classify the video shot as a full-pitched, good-length, or short-pitched delivery. Deep learning methods are the most reliable solution for this task but they also have several limitations, especially when detecting small objects such as a cricket ball. There are many other challenges which will be discussed later.

After extracting the bounce point on the 2-dimensional image of the pitch, the actual position of the ball needs to be calculated in a 3-dimensional frame. To solve this problem, we used object detection along with approximations and the standards that are followed in different broadcast videos. The pitch of the ball is then classified as a full-pitched delivery, a good-length delivery, or a short-pitched delivery. We propose a model which uses video shots from our work on shot segmentation to directly classify a given delivery into one of these three categories. This currently occurs in international cricket where multiple cameras are mounted at different angles on the boundary. However, we only used the front pitch view from the segmented ball delivery dataset as input and detected the pitch position of the ball without the need for any additional equipment.

## 1.4 Contribution

The number of video datasets available online are comparatively less than the image datasets available online. The most popular video dataset available is YouTube-8M Segments published by Google containing 237,000 human verified labeled videos (Google, 2019). Although there are plenty of cricket videos available online, there is not even a single dataset publicly available containing the video segments from the sports of cricket.

Using our approach, we have compiled Cricket Video Dataset (CricViDS) containing unlabeled video segments from the broadcast videos available on YouTube. The dataset contains 3000 unlabeled video clips containing the ball delivery and the stroke play. These videos are from all the formats of the game including Twenty20 matches, one day internationals and test matche. This will enable researchers to extract semantics directly from the video clips instead of processing whole broadcast footage. In future, we are looking to label this dataset with the type of shot played by the batter. Furthermore, the methodology followed in our research can be applied to other sports as well for activity recognition and to compile different sports video datasets.

## 2. Related Work

The extraction of semantics from broadcast videos in sports is a very popular area of interest. Most cricket-based video analysis occurs in the subcontinent due to the popularity of this sport in the region. Several of these studies relate to the generation of highlights using the whole video of a cricket match whereas others are based on annotating cricket videos by deriving semantics from them.

### 2.1 Extraction of Video Shots for Front Pitch View

(Sharma et al., 2015) present an approach to annotate cricket videos with semantic descriptions on a fine-grain spatio-temporal scale. They follow a two-stage approach. In the first stage, they segmented the video into different scenes using the information available in text commentaries and in the second stage, they classified video shots and different phrases in text commentary into different categories.

Another similar study conducted by (Abburu, 2010) presents a tool for the annotation of cricket videos using text extraction from images and voice in videos. The author proposed a DLER tool to extract the semantics from cricket videos. This tool is a robust approach for text detection, localization, extraction, and reorganization in video frames.

(Kumar et al., 2014) determined the direction of the stroke played by the batsman. They first extracted the video frame of the stroke play by segmenting the videos using CUT and FADE techniques, then they applied multiple techniques to identify the direction of the stroke play (Jiang et al., 1998), (Lefèvre et al., 2003). First, they applied a tree-structure approach for batsman pose estimation which didn't work well, so they followed an optical flow analysis approach proposed by (Horn & Schunck, 1981) which achieved an accuracy of 80%.

Another study by (Pramod Sankar et al., 2006) focuses on the temporal segmentation of videos using a multi-modal approach. They used readily available commentary text of the cricket match to segment the videos into meaningful scenes using the scene level description provided by the commentary. They also presented techniques for the generation of highlights automatically.

(Kolekar & Sengupta, 2006) presented a novel approach to generate event-based highlights instead of excitement-based highlights. They used two levels of abstractions to generate these highlights. Although they didn't specify the procedure for the extraction of video shots from the original video, the extraction of semantics from the video is discussed in detail.

Another robust approach in highlight generation uses the oriented FAST, rotated BRIEF (ORB) method with a detection rate of 98.26% (Ringis & Pooransingh, 2015). This work detects video frames where the bowler runs to deliver the ball, termed the Bowler Run-up Sequence (BRS) in the paper. The average matching time was only 0.059 seconds. Although it is very robust, it only detects the video frame of BRS instead of segmenting the whole video shot. The frame classification is done using the FAST method which relies on pixel color and intensity and hence makes the method vulnerable (Taylor et al., 2009).

Chappidi et al. (2009) also extracted the front pitch view using a hidden Markov model (HMM) to generate highlights in cricket. Their approach is based on the observation that highlights are usually made up of transitions between particular shots. They compared different shots against the HMM model and if they matched, they added the shot to the highlight.

Shukla et al. (2018) conducted a comprehensive study on the generation of highlights using an in-depth approach in which event-based and excitement-based features are used to derive important scenes. They used audio cues, such as sudden intensity in the noise from the crowd to identify key moments in the video. They used a multi-stage approach to generate event-driven as well as excitement-driven highlights of the video. In the first stage, they extracted multiple video shots for different overs, in the second stage, they classified the type of scene in the video shot and in the third stage, they generated event-driven and excitement-driven highlights.

In another study, (Deokar & Ruhi, 2015) investigated video shot detection using both global and local histograms along with optical flow to detect cut and fade in video frames. The detection of boundaries in a video shot, despite being an old problem, remains a challenge due to high-speed motion in cricket. This problem was well addressed in their work. After the detection of video shots, they also classified the video shots.

Similar work on shot segmentation in cricket was conducted by (Narasimhan et al., 2010) in which they tried to automatically summarize events in cricket using a genetic algorithm. They used the hue-histogram approach for the segmentation of a shot which contains both ball delivery and stroke play. They detected the shot using the green color of grass and the pitch color. However, this method was unreliable as it gave too many false positives as the whole ground was covered with grass and the pitch can be green on the top as well. Another factor is the crowd in the background which further adds complexity.

(Premaratne et al., 2018) detected video shots from a cricket video by detecting player dismissal in various scenarios. They also used histograms in combination with audio and video features to detect the dismissals but only managed to achieve a classification rate from 54% to 70%. They used videos from 45 different cricket games to analyze the results.

In their comprehensive work, (Thomas et al., 2017) discussed current applications of computer vision in sports. This includes camera calibration, detection and tracking, player modeling and analyzing the motion of players. They also discussed the impact that computer vision techniques have had on sports. They presented the open issues and current research areas for computer vision applications and discussed the computer vision systems that had already been adopted and used.

All of the related work that includes image classification used very small datasets as input and none of them used more than a single cricket match to train their classifiers which is one of the motivations of our work, that is, to present a comprehensive model with a reasonable dataset.

## 2.2   Ball Tracking and Classification

A large body of research has been conducted on ball tracking, particularly in cricket for umpiring assistance, training of players, and performing different in-game analyses. In one such works, (Arora et al., 2017) developed umpiring assistance and a ball tracking system using a smartphone camera but this work is conducted in a special scenario with the bowler delivering the ball in the nets to the batsman and the smartphone camera fixed at a particular known point. They used the sliding window approach to increase the accuracy of ball tracking.

Computer vision has also been applied in tennis on several occasions. (Yan et al., 2014) conducted research to track a tennis ball without any intervention. They applied different techniques for object tracking in computer vision. They computed the trajectory of the tennis ball using candidate level, tracklet level and path level approaches.

Another similar object tracking technique was proposed by (Liu & Carr, 2014) to detect and track the motion of players in sports. They applied random forests on low-level, mid-level and high-level tracklets. The three-level approach is similar to the one proposed by (Yan et al., 2014).

In another study, ball tracking in cricket is undertaken using pure computer vision methods with seeded growing region algorithm and ball candidate generation (Velammal & Kumar, 2010).

Another patented work involving ball tracking using at least four cameras in a spaced-apart relationship was conducted by (Sherry & Hawkins, 2000). This patent was registered for general use in sports but in the case of cricket, they suggested using at least six highly elevated cameras positioned along the ground boundaries to track the ball so that it is less likely to be obscured by the fielders on the ground. This patented work is currently being used as technology termed Hawk-Eye in international cricket to detect ball trajectories to assist umpires make decisions regarding LBWs.

# 3. Methodology

## 3.1 Data Collection & Pre-Preprocessing

The motivation for our work is to provide researchers with a model to build a large-scale open dataset where semantically annotated labeled datasets are available in sports, including cricket. The dataset should have reliable accuracy if it is to be adopted by researchers worldwide. To achieve this objective, our proposed model should have real-time performance so that it can be applied to a broad dataset. The dataset in our case comprises live broadcast videos which are also easily accessible. As our work deals with cricket as a test case, we require a large dataset of broadcast videos. Unfortunately, not many broadcast videos are freely available online. We acquired the broadcast videos of different cricket matches including one-day international matches as well as test matches from YouTube. The videos were from different broadcasters and had different parameters. Their attributes were also of various types including encoding, resolution, frames per second and format of the video. The videos contain both live broadcasted videos as well as videos from the highlights of different cricket matches. The reason for selecting videos in different formats is to ensure the reliability of the model in all circumstances. Furthermore, we also extracted full match videos of the Twenty20 (T20) matches from the Australian T20 league, named the Big Bash League and the Pakistani T20 league named the Pakistan Super League. The videos are readily available on YouTube. These videos also contain general television commercials which increase the diversity of our input dataset. We used free tools available on the Internet to download the videos. The main difference between the shorter formats of the game and the test matches is the color of the players' kits and the color of the ball. The shorter formats of the game include T20 matches and one-day international matches which look similar in the videos apart from a slight change in the playing rules.

## 3.2 Detection of the Front View of the Pitch or Bowler Run-up

A large body of research has been conducted on extracting semantics from broadcast videos to gain useful information in real time. Numerous hurdles had to be overcome to achieve this task, including continuous motion, changes in illumination, partial occlusion, etc. The main problem is that the broadcast videos we deal with involve moving cameras which further increases the complexity of the problem.

To extract the video shots which contain ball deliveries, we first detected the video shot in which the bowler is running to deliver the ball. This video shot is always from the front camera where the camera is facing the batsman directly in the center with the umpire in the same line. The bowler is usually running towards the pitch with the ball in hand and his back towards the camera. The runner stands on the other side of the umpire from which the bowler is delivering the ball. The keeper usually stands behind the batsman and is not always visible from the front view. Some of the camera angles that are used in a broadcast footage are shown in Figure 3.

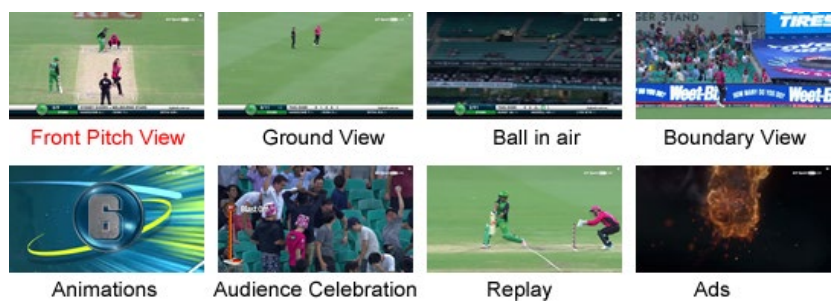

Figure 3. Different types of views in a cricket broadcast

### 3.2.1 Image Classification using MobileNet

We used image classification using MobileNet to classify different frames from the video. In an image classification problem, a model is trained using a training dataset and is tested using a separate test dataset to analyze the efficiency of the model. This is a very common approach used in deep learning where the image must be classified into one of the specified types.

We used ML.net (Microsoft, 2018a) which is an open-source and cross-platform machine learning framework for different machine learning tasks, which includes sentiment analysis, product recommendation, price prediction, customer segmentation, object detection, fraud detection, image classification, etc. Instead of manually identifying hundreds of front pitch view images and negative images, we adopted a transfer learning approach. Transfer learning is a technique used in machine learning to speed up learning using existing knowledge from previously trained models.

To apply transfer learning, we chose the most robust existing model which achieves great accuracy in classifying a variety of images, hence it aligns with our aim to achieve real-time performance. In a study by (Bianco et al., 2018), they presented an in-depth analysis of a majority of DNN models which were currently being used. They analyzed a list of deep learning models for recognition accuracy, model complexity, computational time, memory usage and inference time. These analyses were performed on NVIDIA Titan Xp GPU with Pascal architecture.

We used MobileNet-v2 (Howard et al., 2017) as the best model for our scenario keeping in view its high performance which is shown in terms of FPS. MobileNet-v2 has a total of 53 layers which is significantly fewer compared to the other available models. The pre-trained MobileNet-v2 model was then used along with our dataset to train our model. The interesting thing about this approach is that we were able to train our model in only 40 seconds using a total of 385 images as shown in Table 1.

Table 1. Details for the model trained using transfer learning

| Model Details | |
|---|---|
| Pre-trained Model | MobileNet-v2 |
| Epoch | 50 |
| Batch Size | 10 |
| Learning Rate | 0.01 |
| Total Images | 385 |
| Training Dataset | 80% |
| Test Dataset | 20% |
| Training Time | 40 seconds |

The architecture of our transfer learning approach is shown in Figure 4.

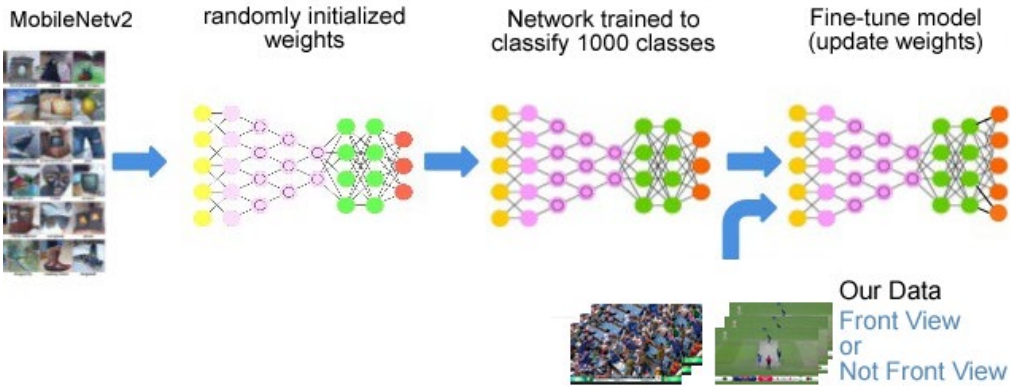

Figure 4. Architecture used to build the model using transfer learning

Our model achieves considerable success which enables our approach to be used in other sports as well. We used a very small dataset of only 385 images to leverage the full potential of transfer learning. The small dataset ensures ease of training and the ability to use the model in other applications in cricket and other sports. Our model was able to achieve 100% accuracy when using a small dataset from a single game which is similar to most of the related works.

### 3.2.2 Umpire and Pitch Annotation Approach Using YOLO

With the advancements in deep learning, it is being applied in a variety of domains including cricket. This includes regression, image classification, and object detection. To classify the front pitch view and other images, we followed a simple object detection approach to annotate the pitches and the back view of the umpire. We trained the annotated dataset using the available object detection models.

We compared different available frameworks for their mean average precision (mAP) and speed (FPS) and chose the YOLO (You Only Look Once) framework as the one best suited to our scenario. The difference in the frameworks was compared on the PASCAL VOC 2007 dataset by (Redmon & Farhadi, 2017).

YOLO is an algorithm that can be implemented using different neural network frameworks, such as PyTorch, Keras or Darknet. Keeping in view one of our primary performance goals, we selected the Darknet implementation of YOLO as it is written in C language and CUDA (NVIDIA, 2007). CUDA is a computing platform developed by NVIDIA for parallel programming over GPU. It allows a substantial increase in computing performance by harnessing the power of the graphics processing unit (GPU).

We used the Alturos.Yolo library for our work which is an open-source darknet wrapper of YOLO on GitHub (AlturosDestinations, 2018). It supports both CPU and GPU to facilitate our aim of developing a flexible model for mass adoption. To annotate the images, we used another open-source software Alturos.ImageAnnotation by the same author to annotate our dataset (AlturosDestinations, 2019). With the aim of ensuring the high reliability of our model, we annotated pitches as well as umpires in the front view of the bowler run-up. The umpire is always standing with his back in front of the bowler run-up scene. We identified the front view of the pitch by detecting these objects in an image to classify it as either the front view of the pitch or other, as shown in Figure 5.

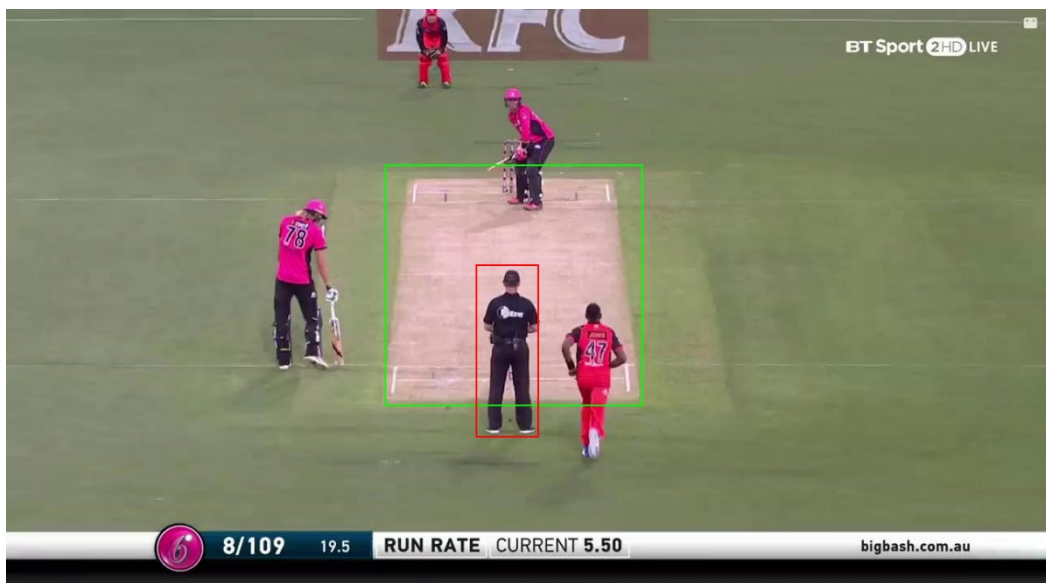

Figure 5. Pitch is annotated with a green rectangle and the umpire is annotated with a red rectangle

The weights were first initialized to be as the default YOLO weights that are available on their website and were then trained on a local machine as the computational requirements were not huge due to limited number of training classes. We used NVIDIA GeForce GTX 1050 GPU (4GB) to train our weights for two hours up to a maximum number of iterations until the weights saturate and the average loss reached minimum and stabilized.

### 3.2.3 Dual-Stage Approach to Increase Precision

Both our image classification and object detection approaches were able to achieve a high recall value but there was room for improvement in terms of precision. To rectify the low value of precision, we converted our model into a dual-stage model where image classification and object detection is undertaken simultaneously. Our core aim is to increase the precision value which occurs at the cost of a reasonable recall value.

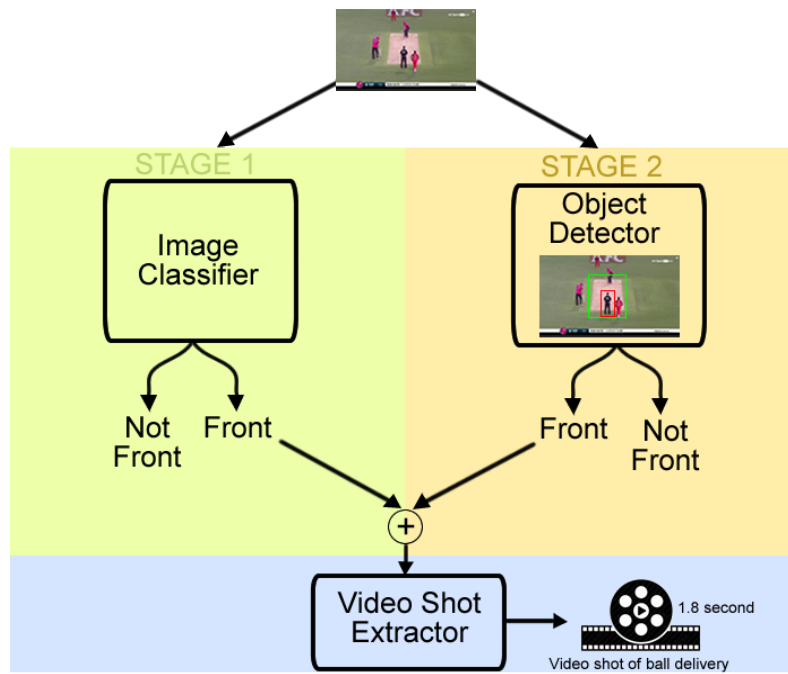

Figure 6. Dual-stage flow for front pitch view detection and shot extraction

We achieved high precision after applying the dual-stage for object detection at the cost of a little processing time incurred by the double processing. The parallel processing of the classifier and object detector is done to keep the performance high by utilizing the parallel processing ability of the GPU. The architecture of the dual-stage pitch view detection and shot extraction is shown in Figure 6.

## 3.3 Replay Identification

After the front pitch view was successfully identified, we categorized the video shots as either replays or live ball deliveries. In cricket, the scorecard at the bottom is only shown when the footage being displayed is live. This scorecard disappears when the replay of the delivery is being shown. The presence of a scorecard is detected when there is a static frame at the bottom using the absolute difference between the first and the last frame of the video shot. The scorecard is usually stationary and doesn't move with the bowler run-up. We applied our method to approximately 1446 video shots for ball deliveries from different formats of the game and it was able to detect replays from all of them. The main aim of scorecard detection is to remove duplicates and include only relevant video shots in the final output.

## 3.4 Video Shot Boundary Detection

The video shot is the sequence of continuous frames until the scene changes. After the front pitch view was detected, the subsequent frames belonging to the same video shot until the end boundary of the video shot were detected. We applied the BackgroundSubtractor class of the OpenCV library to calculate the foreground mask of the frame. After initializing the subtraction model for the first few frames, we calculated the total foreground percentage of the frame. If the foreground percentage of the frame is greater than the threshold, the video shot change is detected. This approach worked very reliably and detected the shot change successfully in 1446 video shots that we used from all formats of the game including videos for the highlights. The foreground mask difference between the consecutive frames at the shot boundary is shown in Figure 7.

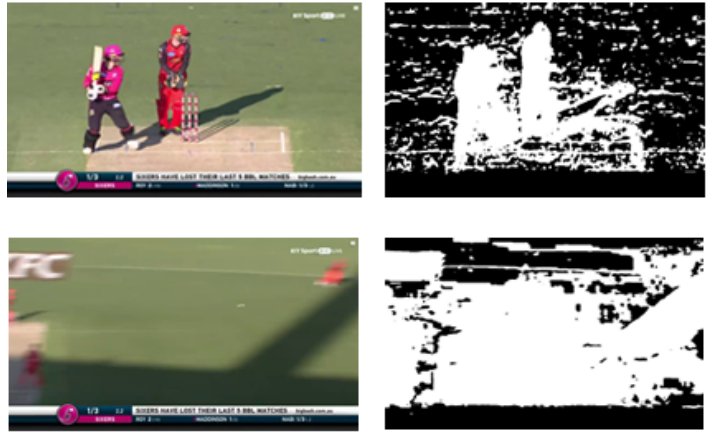
Figure 7. Foreground mask for consecutive frames at the boundary of the video shot

## 3.5 Ball Tracking and Ball Bounce Position Detection

After proposing a model to generate a large reliable dataset of video shots of ball deliveries in cricket, we used our results from the previous section as the input to track the ball using object detection and to also detect the position on the pitch where the ball bounces to classify the delivery as a full-pitched delivery, a good-length delivery or a short-pitched delivery. As evaluated by (Nguyen et al., 2020), the mAP of YOLO for small object detection is quite low. (Lin et al., 2017) proposed an object detector called RetinaNet. RetinaNet is a one-stage object detector similar to YOLO which works very well with small objects. Being a one-stage object detector, the inference speed of RetinaNet is very high but it also surpasses most of the two-stage detectors in terms of accuracy.

The main problem with ball tracking is that a cricket ball moves very fast when it is thrown by the bowler towards the batsman. The ball usually moves as fast as 90 miles per hour. If the shutter speed of the camera or frames per second is not very high, the ball appears as a blurry object, so these images are unreliable for training. To rectify this problem, we extracted images from videos with 50 frames per second. As a test case, we only extracted frames from videos of test cricket because the color of the balls for test cricket is red and for other formats of the game, the color of the ball is white. We extracted 9,232 frames from different videos and annotated 2,255 images containing balls and used the rest of the 6,977 images as the negative images containing no balls. We used VoTT, an open-source annotation tool from Microsoft to annotate the images (Microsoft, 2018b). The complete details of our dataset are shown in Table 2.

Table 2. Dataset details for ball tagging

| Dataset Details | |
|---|---|
| Format of the game | Test Cricket |
| Color of the ball | Red |
| Images Extracted | 9,232 |
| Images annotated with cricket ball | 2,255 |
| Images taken with no ball present | 6,977 |
| Cropped percentage (top, bottom, left, right) | (20%,25%,30%,30%) |

As the images were extracted from 50 FPS videos, the balls were very clear in the images. Using our in-game knowledge, we cropped the extracted images to filter out additional parts of the image as the cricket balls always remain in the focus and are mostly at the center of the screen and can be annotated easily.

We used Python for the Keras implementation of the RetinaNet algorithm to train our images on Google Colab as locally available GPU, NVIDIA GeForce GTX 1050 (4 GB) was not enough to fit our dataset. Google Colab is a product developed by Google that provides the use of free GPU for machine learning and data analysis for research purposes.

We achieved more than 90% accuracy in detecting the ball. The false positives in the continuous frames were eliminated using the shortest-distance approach. The object that is least distant from the position of the ball in the previous frame is more entitled to be classified as a ball. We successfully tracked the ball path in a 2-dimensional image as shown in Figure 8.

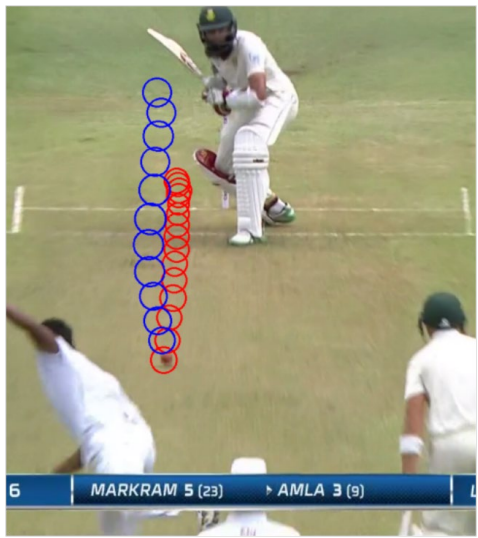

Figure 8. The circles show the 2-dimensional trajectory of the ball on the frame where the ball bounces off the pitch. The red circles show the ball travelling downward whereas the blue circles show the ball travelling going upwards.

The ball is released by the bowler from over his head and is thrown towards the batsman and after bouncing off the pitch, the ball rises upwards towards the batsman. By tracking the 2-dimensional trajectory of the ball on a 2-dimensional image, the position where the ball bounces off the pitch can be easily identified by grabbing the lowest position of the ball on the 2-dimensional image. The 2-dimensional position needs to be converted into a 3-dimensional position to obtain the distance of the ball from the batsman's stumps. This distance classifies the ball into full-pitched, good-length, or short-pitched delivery.

To extract the distance of the ball from the camera or the stumps, the depth of the ball needs to be determined which is usually either done using special depth cameras or fixed cameras from the side of the pitch at the boundary. We applied object detection to determine the approximate distance of the ball from the pitch using the same broadcast footage. To grab the bounce position of the ball on the pitch, the dimensions of the pitch and the camera angle should be known from which the broadcast footage was shot. This was done using approximation and standards followed in cricket.

Although there is no fixed value that defines the range of a good length delivery, most cricketers agree that the ball pitching somewhere between six to eight meters from the stumps is a good-length delivery. This is the length where the batsman is uncertain as to whether to play the ball on the front foot or the back foot. Also, a slight deviation at this length is very difficult to play as the batsman has to judge lateral movement as well as the bounce of the ball and also has less time to respond to it. We used the criteria which is shown in Table 3 to segregate different types of deliveries.

Table 3. Dataset details for categorizing delivery type

| Delivery Type | Pitching distance from batsman's stumps |
|---|---|
| Full-Pitched Delivery | Less than 6 meters |
| Good-Length Delivery | 6 to 8 meters (Some people consider 6 to 9 meters as good-length) |
| Short-Pitched Delivery | More than 8 meters |

Using the above criteria, we designed the figure of the pitch from the bowler's crease to the batsman's crease which is scaled according to the dimensions of the pitch excluding the distance from stumps to crease as shown in Figure 9.

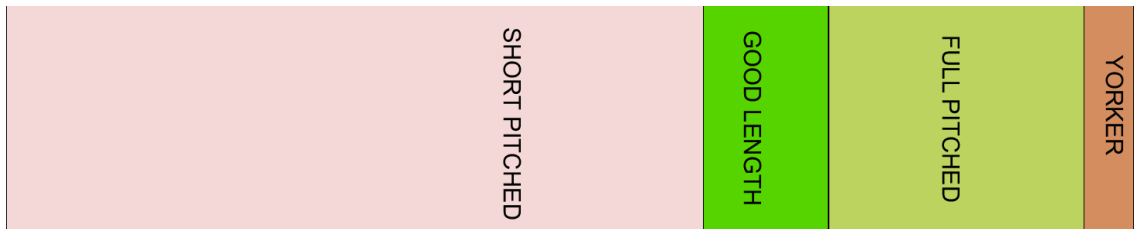
Figure 9. Distribution of different types of deliveries on a cricket pitch (to-scale)

We used this distribution image and mapped it on the pitch vertically from the batsman's crease to the bowler's crease. To take into account the angle that the broadcast camera has with the pitch, we used the trial and error method. We determined that by rotating the image 20 degrees on the y-axis from the front perspective, we were able to match the actual pitch in most cases. The distribution image mapped to the actual image is shown in Figure 10.

As the ball travels towards the batsman after being released by the bowler, the front camera continues to be zoomed on the batsman to give a clear view of the batsman and the ball. There is no marker on the pitch to calculate how much the camera has zoomed into the batsman. We applied ResNet50 model to calculate this zoom factor. We used the frame when the ball is released from the bowler's hand and the frame when it bounces off the pitch to calculate the zoom factor and the position where the ball bounces off the pitch.

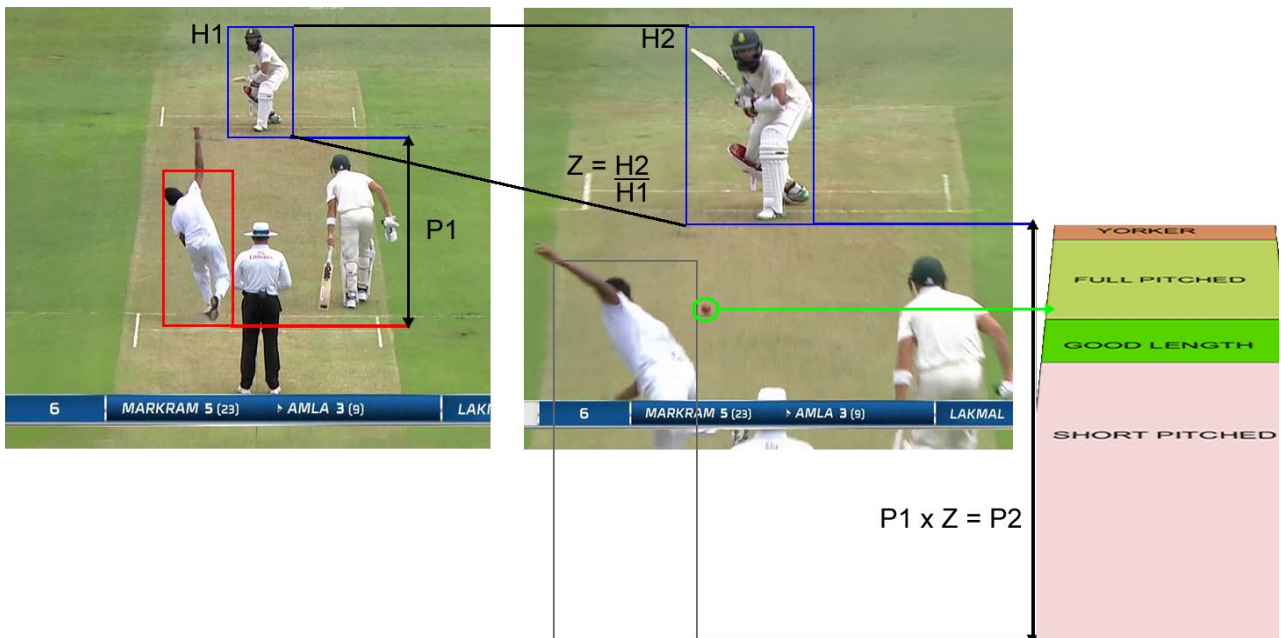
Figure 10. Calculation of the bounce position of the ball

The following steps are used to calculate the bounce position of the ball:

*Step 1:*
Detect the batsman and the bowler using the ResNet50 model and calculate the height of the batsman.

Let H1 be the height of the batsman in the first frame where the bowler releases the ball.

*Step 2:*
Calculate the height of the pitch from the bottom of the batsman to the bottom of the bowler in the same frame.

Let P1 be the height of the pitch in the first frame.

*Step 3:*
Detect the batsman using object detection again in the second frame.

Let H2 be the height of the batsman in the second frame where the bowler releases the ball.

*Step 4:*
Calculate the zoom factor of the second frame using the following formula:
$$Z = \frac{H2}{H1}$$

*Step 5:*
Calculate the height of the full pitch in the second frame using the following formula:
$$P2 = Z * P1$$

*Step 6:*
After the height of the pitch in the second frame is known, map the distribution image on the pitch and categorize the type of ball by checking the position where the ball bounces off the surface.

As shown in Figure 10, the type of ball is not discrete but continuous. The ball in Figure 10 pitches just above the good-length delivery which is still considered to be a good length by most viewers. The details of the dataset we used for classifying our dataset are shown in Table 4.

Table 4. Dataset details for classification of delivery type

| Dataset Details | |
|---|---|
| Format of the game | Test Cricket |
| Color of the ball | Red |
| Deliveries Categorized | 214 |
| Model used for object detection (bowler and batsman) | ResNet50 |
| Model used for ball detection | RetinaNet using Keras |

# 4. Results

## 4.1 Detection of the front view of the pitch or bowler run-up

We discuss the results for the different approaches we used to detect the front view of the pitch or the scene where the bowler runs to deliver the ball.

### *4.1.1 Image Classification using MobileNet*

To test the prediction accuracy of the model, we compiled a comprehensive dataset of 497,040 images containing an equal number of the front pitch view and other images from several different videos which include T20 matches, one-day international matches, and test matches (Reference for the dataset?). The confusion matrix for the result is shown in Table 5.

Table 5. Confusion matrix for the results using transfer learning

| n=497,040 | Predicted: FRONT | Predicted: NOT FRONT | |
|---|---|---|---|
| Actual: FRONT | TP = 233,358 | FN = 15,162 | 248,520 |
| Actual: NOT FRONT | FP = 4,845 | TN = 243,675 | 248,520 |
| | 238,203 | 258,837 | |

The confusion matrix above shows remarkable results considering that a small training dataset of only 385 images was used. The recall and precision for our model are shown as follows.

$$\text{Recall} = \frac{TP}{TP+FN} = \frac{233,358}{233,358+15,162} = 93.89\%$$

$$\text{Precision} = \frac{TP}{TP+FP} = \frac{233,358}{233,358+4,845} = 97.96\%$$

High values of recall and precision make our model easily adoptable in a large number of scenarios. The same approach can be applied to different sports other than cricket.

Most live sports broadcasts are shot at 30 frames per second. Our model surpassed this by 9 frames per second on a low-end GPU which enables our model to be easily adopted in live broadcast scenarios. We used EmguCV which is a .NET wrapper for computer vision library OpenCV to grab frames from the video and perform different tasks (Emgu Corporation, 2015). The performance results for our model are shown in the Table 6.

Table 6. Performance results using transfer learning

| Performance Results | |
|---|---|
| Images Predicted | 497,040 frames from different videos |
| Prediction time per frame | 25.56 ms |
| Prediction Speed (FPS) | 39 frames per second |
| Overhead for grabbing one frame from video | 10.17 ms |

### 4.1.2 Umpire and Pitch annotation approach using YOLO

The model that we developed by annotating the pitch was then tested on the same testing dataset which was used to test the image classification approach detailed in the previous section. The confusion matrix for the detection of the front view using the umpire and pitch is shown separately in Tables 7 and 8, respectively.

Table 7. Confusion matrix for the results using **object detection of the umpire's back view**

| n=497,040 | Predicted: FRONT | Predicted: NOT FRONT | |
|---|---|---|---|
| **Actual: FRONT** | TP = 209,532 | FN = 38,988 | 248,520 |
| **Actual: NOT FRONT** | FP = 14,022 | TN = 234,498 | 248,520 |
| | 223,554 | 273,486 | |

The recall and precision for front pitch view detection using object detection of the umpire's back view did not give better results for recall and precision compared to image classification using the transfer learning approach shown as follows:

$$\text{Recall} = \frac{TP}{TP+FN} = \frac{209,532}{209,532+38,988} = 84.31\%$$

$$\text{Precision} = \frac{TP}{TP+FP} = \frac{209,532}{209,532+14,022} = 93.72\%$$

The results using the same approach but for direct pitch detection are given in Table 8.

Table 8. Confusion matrix for the results using **object detection of pitch**

| n=497,040 | Predicted: FRONT | Predicted: NOT FRONT | |
|---|---|---|---|
| **Actual: FRONT** | TP = 240,084 | FN = 8,436 | 248,520 |
| **Actual: NOT FRONT** | FP = 12,255 | TN = 236,265 | 248,520 |
| | 252,339 | 244,701 | |

This approach to detect annotated pitches in the images directly to identify the front pitch view gives better results for both recall and precision compared to umpire detection and image classification. The precision for this approach is comparable to image classification using the transfer learning approach.

$$\text{Recall} = \frac{TP}{TP+FN} = \frac{240,084}{240,084+8,436} = 96.60\%$$

$$\text{Precision} = \frac{TP}{TP+FP} = \frac{240,084}{240,084+12,255} = 95.14\%$$

We also calculated the results for both of the objects (umpire and pitch) to detect the front pitch view so that if either of the objects is detected, we considered it as a front pitch view. The results for the approach are shown in the confusion matrix in Table 9.

Table 9. Confusion matrix for the results using **object detection of either umpire or pitch**

| n=497,040 | **Predicted: FRONT** | **Predicted: NOT FRONT** | |
|---|---|---|---|
| **Actual: FRONT** | TP = 241,452 | FN = 7,068 | 248,520 |
| **Actual: NOT FRONT** | FP = 17,841 | TN = 230,679 | 248,520 |
| | 259,293 | 237,747 | |

By using this approach, the recall value is increased by a reasonable amount.

$$\text{Recall} = \frac{TP}{TP+FN} = \frac{241,452}{241,452+7,068} = 97.15\%$$

$$\text{Precision} = \frac{TP}{TP+FP} = \frac{241,452}{241,452+17,841} = 93.11\%$$

Finally, the results for the dual-stage model where image classification and object detection is undertaken simultaneously are shown in Table 10.

Table 10. Confusion matrix for the **dual-stage model** using image classification as well as object detection

| n=497,040 | **Predicted: FRONT** | **Predicted: NOT FRONT** | |
|---|---|---|---|
| **Actual: FRONT** | TP = 248,349 | FN = 171 | 248,520 |
| **Actual: NOT FRONT** | FP = 45,942 | TN = 202,578 | 248,520 |
| | 294,291 | 202,749 | |

By using this approach, we achieved a high value for recall. The major drawback of this approach is the low value of precision, shown as follows:

$$\text{Recall} = \frac{TP}{TP+FN} = \frac{248,349}{248,349+171} = 99.93\%$$

$$\text{Precision} = \frac{TP}{TP+FP} = \frac{248,349}{248,349+45,942} = 84.38\%$$

The recall and precision for different methods are shown in Figure 11.

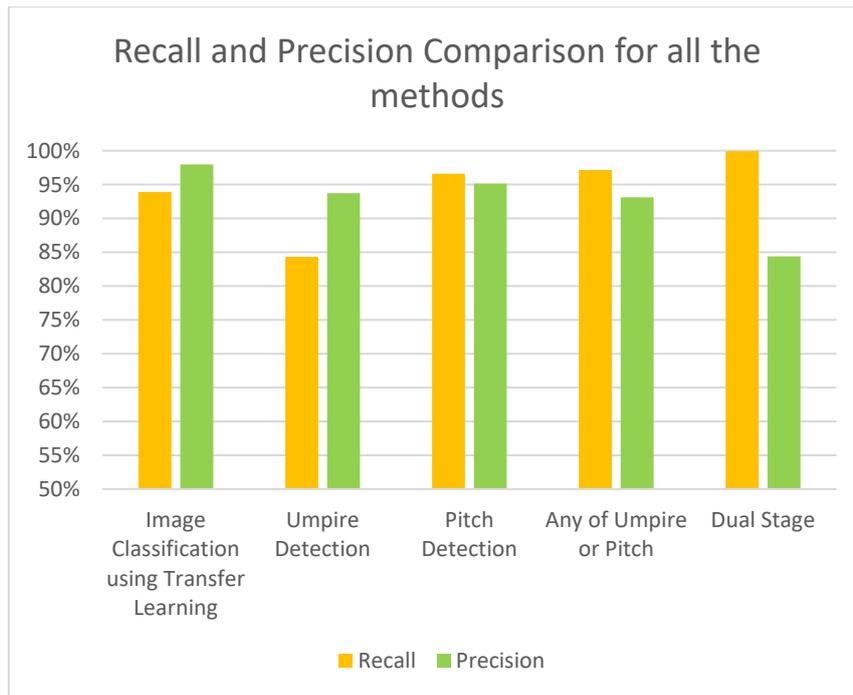

Figure 11. Recall and precision for all the methods

Our single-stage object detection models achieved 312 frames per second. This robustness of our model enables it to be used in scenarios where very high performance is required. The performance of our model is still underrated as the results are not from the top-performing GPUs in the market. The performance results for our model are shown in Table 11.

Table 11. Performance results using object detection

| **Performance Results** | |
|---|---|
| Images Predicted | 497,040 frames from different videos |
| Prediction time per frame | 3.2 ms |
| Prediction Speed (FPS) | 312 frames per second |
| GPU | NVIDIA GeForce GTX 1050 (4GB) |

## 4.2 Ball Bounce Detection

The position of the pitch where the ball bounced is calculated to classify the different delivery types. Although all the results looked correct from the viewer's perspective, there is no concrete yardstick to test the results again, as a good-length delivery for one viewer can be a full-pitched delivery for another. The results are shown in Table 12.

Table 12. Results for Ball Bounce Detection

| **Ball Bounce Detection Results** | |
|---|---|
| Format of the game | Test Cricket |
| Color of the ball | Red |
| Deliveries Categorized | 214 |
| Full-Pitched Deliveries | 80 |
| Good-Length Deliveries | 85 |
| Short-Pitched Deliveries | 49 |

# 5. Conclusion

Our work proposes a novel method to solve a largely unaddressed problem in cricket. When video shots are concatenated in a single video, our model was able to compress the original broadcast video twenty times which can be very helpful for cricket coaches and players. Using the proposed method, researchers can easily construct a large dataset of cricket video shots from original cricket broadcasts.

The proposed dual-stage model achieved 99.93% precision which makes our research viable for commercial use as well. Also, achieving more than 93% for both precision and recall by training only 385 images in 40 seconds shows the potential of using the transfer learning approach in different sports for the extraction of semantics.

We intend to extend our work to further detect the type of shot played by the batsman and then propose a model to automatically generate cricket commentary using generative adversarial networks (GAN). The aim of using GAN is to propose a model which outperforms humans in terms of performance and the quality of cricket commentary.

# 6. Availability of Data

Readers can find the sample processed output dataset of the front camera view of Cricket generated using the model outlined in this manuscript. We have chosen to provide intermediary output images samples of extracted front view to a) avoid copyright issues which can arise from releasing unprocessed raw video, and b) to make it easy for anyone who wants to further develop segmentation of ball deliveries. This will relieve the researchers and practitioners from extracting the most important part and view of the game of Cricket (the front view). Extraction of this part from raw video is a non-trivial task. The dataset can be downloaded from: https://github.com/theimad/Cricket-Image-Segmentation

# REFERENCES

Abburu, S. (2010). Semantic Segmentation and Event Detection in Sports Video using Rule Based Approach. *International Journal of Computer Science and …*, *10*(10), 35–40.

AlturosDestinations. (2018). *Alturos.Yolo, Realtime object detection systems*. https://github.com/AlturosDestinations/Alturos.Yolo

AlturosDestinations. (2019). *Alturos.ImageAnnotation, Open source object annotation software*. https://github.com/AlturosDestinations/Alturos.ImageAnnotation

Arora, U., Verma, S., Sahni, S., & Sharma, T. (2017). Cricket umpire assistance and ball tracking system using a single smartphone camera. *PeerJ PrePrints*, *5*, 1–14. https://doi.org/10.7287/peerj.preprints.3402

Bianco, S., Cadene, R., Celona, L., & Napoletano, P. (2018). Benchmark analysis of representative deep neural network architectures. *IEEE Access*, *6*, 64270–64277. https://doi.org/10.1109/ACCESS.2018.2877890

Deokar, M., & Ruhi, P. (2015). *Video Shot Detection & Classification in Cricket Videos*. *4*(7).

Emgu Corporation. (2015). *EmguCV, Open source and cross-platform wrapper for OpenCV library*.

Google. (2019, June). *YouTube-8M Segments*. https://research.google.com/youtube8m/explore.html

Gupta, A., & Sakthi Balan, M. (2019). Cricket stroke extraction: Towards creation of a large-scale cricket actions dataset. *ArXiv*.

Horn, B. K. P., & Schunck, B. G. (1981). Determining optical flow. *Artificial Intelligence*, *17*(1), 185–203. https://doi.org/https://doi.org/10.1016/0004-3702(81)90024-2

Howard, A., Zhu, M., Chen, B., Kalenichenko, D., Wang, W., Weyand, T., Andreetto, M., & Adam, H. (2017). *MobileNets: Efficient Convolutional Neural Networks for Mobile Vision Applications*.

Jiang, H., Helal, A. (Sumi), Elmagarmid, A. K., & Joshi, A. (1998). Scene change detection techniques for video database systems. *Multimedia Systems*, *6*(3), 186–195. https://doi.org/10.1007/s005300050087

Kobla, V., DeMenthon, D., & Doermann, D. (1999). Detection of slow-motion replay sequences for identifying sports videos. *1999 IEEE Third Workshop on Multimedia Signal Processing (Cat. No.99TH8451)*, 135–140. https://doi.org/10.1109/mmsp.1999.793810

Kolekar, M. H., & Sengupta, S. (2006). Event-Importance Based Customized and Automatic Cricket Highlight Generation. *2006 IEEE International Conference on Multimedia and Expo*, 1617–1620. https://doi.org/10.1109/ICME.2006.262856

Kumar, A., Garg, J., & Mukerjee, A. (2014). Cricket activity detection. *International Image Processing, Applications and Systems Conference, IPAS 2014*. https://doi.org/10.1109/IPAS.2014.7043264

Lefèvre, S., Holler, J., & Vincent, N. (2003). A review of real-time segmentation of uncompressed video sequences for content-based search and retrieval. *Real-Time Imaging*, *9*(1), 73–98. https://doi.org/https://doi.org/10.1016/S1077-2014(02)00115-8

Lin, T., Goyal, P., Girshick, R., He, K., & Dollár, P. (2017). Focal Loss for Dense Object Detection. *2017 IEEE International Conference on Computer Vision (ICCV)*, 2999–3007. https://doi.org/10.1109/ICCV.2017.324

Liu, J., & Carr, P. (2014). *Detecting and Tracking Sports Players with Random Forests and Context-Conditioned Motion Models BT - Computer Vision in Sports* (T. B. Moeslund, G. Thomas, & A. Hilton, Eds.; pp. 113–132). Springer International Publishing. https://doi.org/10.1007/978-3-319-09396-3_6

Microsoft. (2018a). *ML.NET, Open source and cross-platform machine learning framework*.

Microsoft. (2018b). *VoTT, Open-source annotation tool for machine learning*. https://github.com/microsoft/VoTT

Narasimhan, H., Satheesh, S., & Sriram, D. (2010). Automatic summarization of cricket video events using genetic algorithm. *Proceedings of the 12th Annual Genetic and Evolutionary Computation Conference, GECCO '10 - Companion Publication*, 2051–2054. https://doi.org/10.1145/1830761.1830858

Nguyen, N. D., Do, T., Ngo, T. D., & Le, D. D. (2020). An Evaluation of Deep Learning Methods for Small Object Detection. *Journal of Electrical and Computer Engineering*. https://doi.org/10.1155/2020/3189691

NVIDIA. (2007). *NVIDIA CUDA Technology, Parallel computing platform and programming model*.

Pramod Sankar, K., Pandey, S., & Jawahar, C. V. (2006). *Text Driven Temporal Segmentation of Cricket Videos*. 433–444. https://doi.org/10.1007/11949619_39

Premaratne, S. C., Jayaratne, K. L., & Sellapan, P. (2018). Improving event resolution in cricket videos. *ACM International Conference Proceeding Series*, 69–73. https://doi.org/10.1145/3282286.3282293

Redmon, J., & Farhadi, A. (2017). YOLO9000: Better, faster, stronger. *Proceedings - 30th IEEE Conference on Computer Vision and Pattern Recognition, CVPR 2017*, 6517–6525. https://doi.org/10.1109/CVPR.2017.690

Ringis, D., & Pooransingh, A. (2015). Automated highlight generation from cricket broadcasts using ORB. *2015 IEEE Pacific Rim Conference on Communications, Computers and Signal Processing (PACRIM)*, 58–63. https://doi.org/10.1109/PACRIM.2015.7334809

Sharma, R. A., Pramod Sankar, K., & Jawahar, C. V. (2015). Fine-grain annotation of cricket videos. *Proceedings - 3rd IAPR Asian Conference on Pattern Recognition, ACPR 2015*, 421–425. https://doi.org/10.1109/ACPR.2015.7486538

Sherry, D., & Hawkins, P. (2000). *VIDEO PROCESSOR SYSTEMS FOR BALL TRACKING IN GAMES* (Patent No. WO 01/41884 A1). World Intellectual Property Organization, International Bureau.

Taylor, S., Rosten, E., & Drummond, T. (2009). Robust feature matching in 2.3μs. *2009 IEEE Computer Society Conference on Computer Vision and Pattern Recognition Workshops*, 15–22. https://doi.org/10.1109/CVPRW.2009.5204314

Thomas, G., Gade, R., Moeslund, T. B., Carr, P., & Hilton, A. (2017). Computer vision for sports: Current applications and research topics. *Computer Vision and Image Understanding*, *159*, 3–18. https://doi.org/10.1016/j.cviu.2017.04.011

Velammal, B. L., & Kumar, P. A. (2010). An Efficient Ball Detection Framework for Cricket. *International Journal of Computer Science Issues*, *7*(3), 30–35.

Yan, F., Christmas, W., & Kittler, J. (2014). *Ball Tracking for Tennis Video Annotation BT - Computer Vision in Sports* (T. B. Moeslund, G. Thomas, & A. Hilton, Eds.; pp. 25–45). Springer International Publishing. https://doi.org/10.1007/978-3-319-09396-3_2